\documentclass{article}
\pdfoutput=1 

\usepackage{microtype}
\usepackage{graphicx}
\usepackage{subfigure}
\usepackage{booktabs} 

\usepackage{multirow}
\usepackage{hyperref}

\usepackage[accepted]{icml2018}

\icmltitlerunning{On representation power of neural network-based graph embedding and beyond}

\usepackage{amsmath,amssymb}
\usepackage{algorithmic,algorithm}
\usepackage{subfigure}
\usepackage{multirow}
\usepackage{comment}
\usepackage{enumerate}
\usepackage{wrapfig}
\usepackage{ascmac}

\usepackage{mdframed}

\usepackage{titling}
\usepackage{natbib}

\usepackage{framed}

\newcommand{\bs}{\boldsymbol}
\newcommand{\x}{\bs x}

\newcommand{\diff}{\mathrm{d}}

\def\qed{\hfill $\Box$} 

\newtheorem{ex}{Example}[section]

\allowdisplaybreaks[1]

\begin{document}

\twocolumn[
\icmltitle{On representation power of neural network-based graph embedding \\ and beyond}



\icmlsetsymbol{equal}{*}

\begin{icmlauthorlist}
\icmlauthor{Akifumi Okuno}{kyoto,riken}
\icmlauthor{Hidetoshi Shimodaira}{kyoto,riken}
\end{icmlauthorlist}

\icmlaffiliation{kyoto}{Graduate School of Informatics, Kyoto University, Kyoto, Japan}
\icmlaffiliation{riken}{RIKEN Center for Advanced Intelligence Project (AIP), Tokyo, Japan}

\icmlcorrespondingauthor{Akifumi Okuno}{okuno@sys.i.kyoto-u.ac.jp}

\icmlkeywords{Feature Learning, Representation theorem, Neural Network, Indefinite kernels, Non-euclidean}

\vskip 0.3in
]



\newtheorem{theo}{Theorem}[section]
\newtheorem{defi}{Definition}[section]
\newtheorem{lemm}{Lemma}[section]
\newtheorem{prop}{Proposition}[section]

\printAffiliationsAndNotice{}  

\begin{abstract}
We consider the representation power of siamese-style similarity functions used in neural network-based graph embedding.
The inner product similarity~(IPS) with feature vectors computed via neural networks
is commonly used for representing the strength of association between two nodes.
However, only a little work has been done on the representation capability of IPS.
A very recent work shed light on the nature of IPS and reveals that IPS has the capability of approximating any positive definite~(PD) similarities. 
However, a simple example demonstrates the fundamental limitation of IPS to approximate non-PD similarities. 
We then propose a novel model named Shifted IPS~(SIPS) that approximates any Conditionally PD~(CPD) similarities arbitrary well.
CPD is a generalization of PD with many examples such as negative Poincar\'e distance and negative Wasserstein distance, thus SIPS has a potential impact to significantly improve the applicability of graph embedding without taking great care in configuring the similarity function.
Our numerical experiments demonstrate the SIPS's superiority over IPS. 
In theory, we further extend SIPS beyond CPD by considering the inner product in Minkowski space so that it approximates more general similarities.
\end{abstract}

\begin{figure}[htbp]
\centering
\subfigure[True]{
		\includegraphics[width=2.5cm]{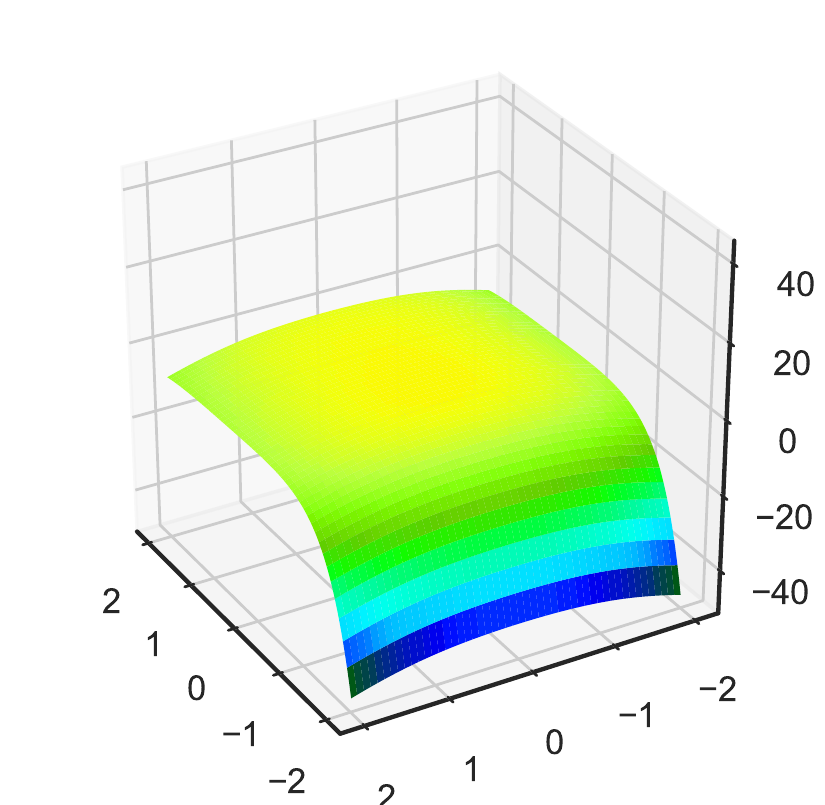}
	\label{fig:true}
}
\subfigure[ Existing~(IPS) ]{
		\includegraphics[width=2.5cm]{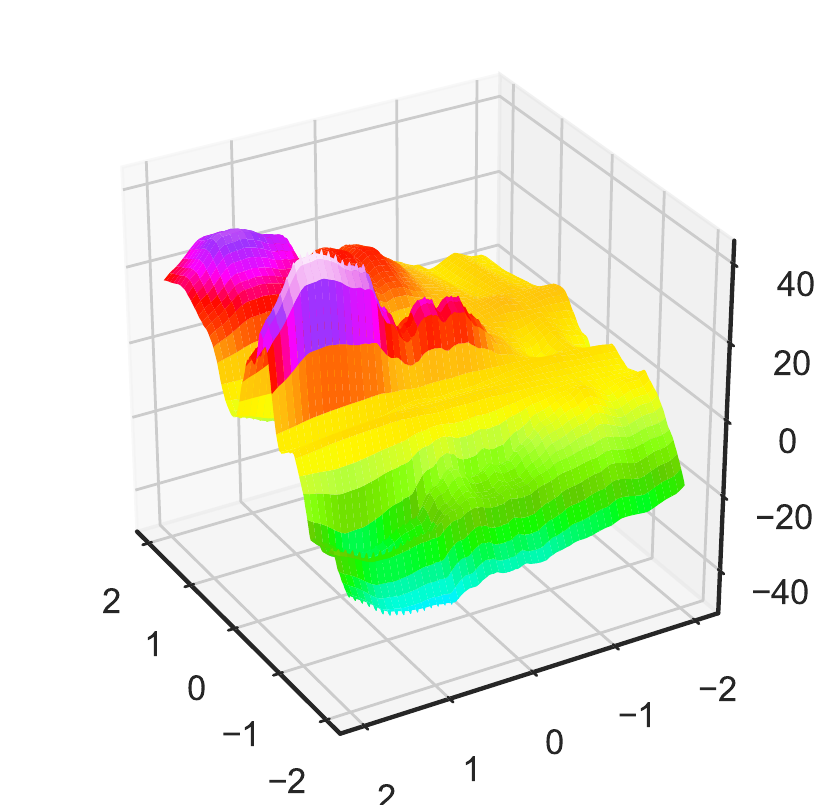}
	\label{fig:existing}
}
\subfigure[ Proposed~(SIPS) ]{
		\includegraphics[width=2.5cm]{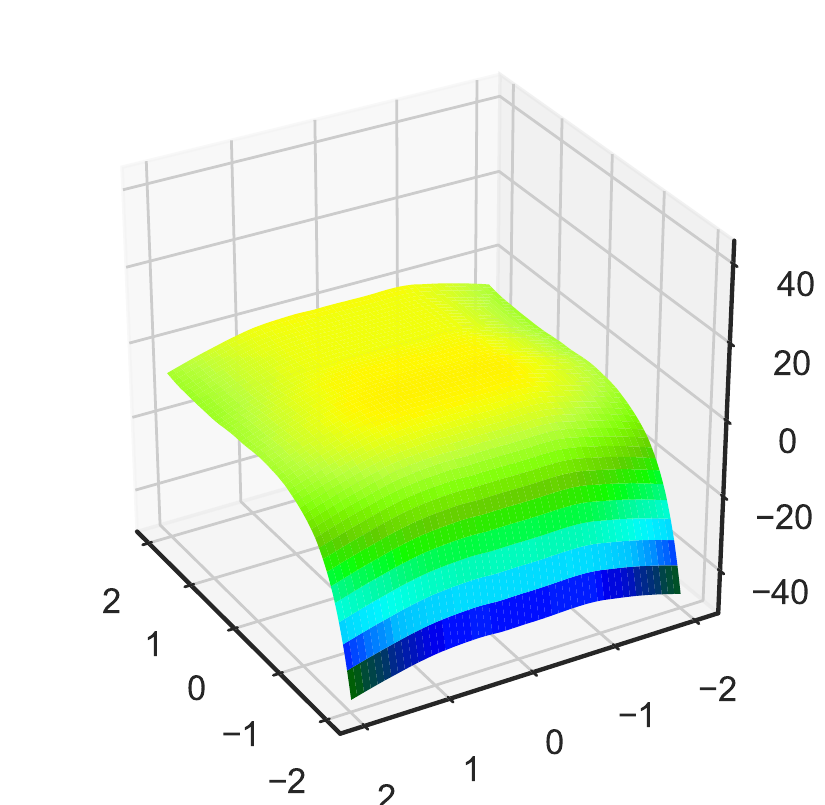}
	\label{fig:proposal1}
}
\caption{\subref{fig:true}~For $f_*(\bs x)=(x_1,\cos x_2,\exp(-x_3),\sin(x_4-x_5)) \in \mathbb{R}^4$ w.r.t. $\bs x\in \mathbb{R}^5$,
the negative squared distance~(NSD) similarity $-\|f_*(s \bs e_1)-f_*(t \bs e_2)\|_2^2$ 
is plotted on $(s,t)$-plane along with two orthogonal directions $\bs e_1,\bs e_2 \in \mathbb{R}^5$.
This NSD is approximated by the two similarity models:
\subref{fig:existing}~Existing model~(IPS) $\langle f_{\bs \psi}(s \bs e_1),f_{\bs \psi}(t \bs e_2) \rangle$,
and 
\subref{fig:proposal1}~Proposed model~(SIPS) $\langle f_{\bs \psi}(s \bs e_1),f_{\bs \psi}(t \bs e_2) \rangle + u_{\bs \xi}(s \bs e_1)+u_{\bs \xi}(t \bs e_2)$,
where $f_{\bs \psi}:\mathbb{R}^{5} \to \mathbb{R}^{10}$ and $u_{\bs \xi}:\mathbb{R}^5 \to \mathbb{R}$ are two-layer neural networks with $1{,}000$ hidden units and ReLU activations. 
The proposed model~(SIPS) approximates the NSD better than the existing model~(IPS). 
}
\label{fig:illustrative_example}
\end{figure}

\section{Introduction}

Graph embedding~(GE) of relational data, such as texts, images, and videos, etc., now plays an indispensable role in machine learning. 
To name but a few, words and contexts in a corpus constitute relational data, and their vector representations obtained by skip-gram model~\citep{mikolov2013distributed} and GloVe~\citep{pennington2014glove} are often used in natural language processing.
More classically, a similarity graph is first constructed from data vectors, and nodes are embedded to a lower dimensional space where connected nodes are closer to each other~\cite{cai2018comprehensive}.

Embedding is often designed so that the inner product between two vector representations in Euclidean space expresses their similarity.
In addition to its interpretability, the inner product similarity has the following two desirable properties:
(1)~The vector representations are suitable for downstream tasks as feature vectors because machine learning methods are often based on inner products (e.g., kernel methods).
(2)~Simple vector arithmetic in the embedded space may represent similarity arithmetic such as
the ``linguistic regularities'' of word vectors \cite{mikolov2013linguistic}.
The latter property comes from the distributive law of inner product
$\langle \bs a + \bs b , \bs c \rangle = \langle \bs a, \bs c\rangle + \langle \bs b , \bs c \rangle$,
which decomposes the similarity of $\bs a + \bs b $ and $\bs c$ into the sum of the two similarities.
For seeking the word vector $\bs y' = \bs y_\text{queen}$, we maximize
$\langle \bs y_\text{king} - \bs y_\text{man} + \bs y_\text{woman}, \bs y' \rangle
= \langle \bs y_\text{king}, \bs y' \rangle - \langle \bs y_\text{man}, \bs y' \rangle + \langle \bs y_\text{woman}, \bs y' \rangle $ in Eq.~(3) of \citet{levy2014linguistic}.
Thus solving analogy questions with vector arithmetic is mathematically equivalent to seeking a word which is similar to king and woman but is different from man.

While classical GE has been quite successful, it considers simply the graph structure, where data vectors (pre-obtained attributes such as color-histograms of images), if any, are used only through the similarity graph.
To fully utilize data vectors, neural networks~(NNs) are incorporated into GE so that data vectors are converted to new vector representations~\citep{kipf2016variational,zhang2017user,hamilton2017inductive,dai2018adversarial}, which reduces to the classical GE by taking 1-hot vectors as the data vectors.
While these methods consider $1$-view setting, multi-view setting is considered in Probabilistic Multi-view Graph Embedding~\citep[PMvGE]{okuno2018probabilistic}, which generalizes existing multivariate analysis methods (e.g., PCA and CCA) and NN-extensions \citep[DCCA]{andrew2013deep} as well as graph embedding methods such as 
Locality Preserving Projections~\citep[LPP]{he2003locality,yan2007graph},
Cross-view Graph Embedding~\citep[CvGE]{huang2012cross}, and
Cross-Domain Matching Correlation Analysis~\citep[CDMCA]{shimodaira2016cross}.
In these methods, the inner product of two vector representations obtained via NNs represents the strength of association between the corresponding two data vectors.
The vector representations and the inner products are referred to as \emph{feature vectors} and \emph{Inner Product Similarities (IPS)}, respectively, in this paper.



IPS is considered to be highly expressive for representing the association between data vectors
due to the Universal Approximation Theorem~\citep[UAT]{funahashi1989approximate,cybenko1989approximation,
yarotsky2016error,pmlr-v70-telgarsky17a}, which proves that NNs having many hidden units approximate arbitrary continuous functions within any given accuracy. 
However, since IPS considers the inner product of two vector-valued NNs, 
the UAT is not directly applicable to the whole network with the constraints at the final layer,
thus its representation capability is yet to be clarified. 
For that reason, \citet{okuno2018probabilistic} incorporates UAT into Mercer's theorem~\citep{minh2006mercer} and proves that IPS approximates any similarity based on Positive Definite~(PD) kernels arbitrary well. 
This result shows not only the validity but also the fundamental limitation of IPS, meaning that the PD-ness of the kernels is required for IPS to approximate. 



To overcome the limitation, similarities based on specific kernels other than inner products have received considerable attention in recent years. 
One example is Poincar\'e embedding~\citep{nickel2017poincare} which is an NN-based GE using Poincar\'e distance for embedding vectors in hyperbolic space instead of Euclidean space.
Hyperbolic space is especially compatible with computing feature vectors of tree-structured relational data~\citep{sarkar2011low}. 
Similarly, Gaussian embedding~\citep{vilnis2014word,bojchevski2018deep} is proposed to learn features based on Kullback-Leibler divergence. 
While these methods efficiently compute reasonable low-dimensional feature vectors by virtue of specific kernels, their theoretical differences from IPS is not well understood.



In order to provide theoretical insights on these methods, in this paper, we first point out that some specific kernels used in above methods are not PD by referring to existing studies. 
To deal with such non-PD kernels, we consider Conditionally PD~(CPD) kernels~\citep{berg1984harmonic,scholkopf2001kernel} which include PD kernels as special cases. 
We then propose a novel model named \emph{Shifted IPS~(SIPS)} that approximates similarities based on CPD kernels within any given accuracy. 
We show an illustrative example in Figure~\ref{fig:illustrative_example}.
Interestingly, negative Poincar\'e distance is already proved to be CPD~\citep{faraut1974distances} and it is not PD. So, similarities based on this kernel can be approximated by SIPS but not by IPS.




Our contribution in this paper is summarized as follows:
\begin{enumerate}[{(1)}]
\item We review existing studies on IPS. Although IPS approximates similarities based on PD kernels arbitrary well, we point out the fundamental limitation of IPS to approximate similarities based on CPD kernels. 
\item We propose a novel model named Shifted IPS~(SIPS) and prove that SIPS approximates similarities based on CPD kernels within any given accuracy. A simper version of SIPS as well as a further extended model beyond CPD is also discussed.
\item We perform a numerical experiment to compare SIPS with IPS. 
\end{enumerate}

The remaining of this paper is organized as follows. 
In Section~\ref{sec:background}, we introduce Inner Product Similarity~(IPS) model, which is commonly used in NN-based GE. 
In Section~\ref{sec:pd_kernels}, we review the previous study on IPS for approximating PD kernels.
In Section~\ref{sec:cpd_kernels}, we show the fundamental limitation of IPS, and then we propose a novel model named SIPS, so that it approximates any similarities based on CPD kernels arbitrary well. 
In Section~\ref{sec:experiment}, we conduct a numerical experiment to compare SIPS with IPS. 
In Section~\ref{sec:conclusion}, we conclude this paper.
In Appendix~\ref{sec:arbitrary_kernels}, we also mention a further extended model based on the inner product in Minkowski space for more general similarities beyond CPD.

\section{Background: Generative model for NN-based feature learning}
\label{sec:background}
We consider an undirected graph consisting of $n$ nodes $\{v_i\}_{i=1}^{n}$ and link weights $\{w_{ij}\}_{i,j=1}^{n} \subset \mathbb{R}_{\geq 0}$ satisfying $w_{ij}=w_{ji}$ and $w_{ii}=0$, where
$w_{ij}$ represents the strength of association between $v_i$ and $v_j$.
The data vector representing the attributes (or side-information) at $v_i$ is denoted as $\bs x_i \in \mathbb{R}^p$.
If we have no attributes, we use 1-hot vectors in $\mathbb{R}^n$ instead. 
We assume that  $\{w_{ij}\}_{i,j=1}^{n},\{\bs x_i\}_{i=1}^{n}$ are obtained as observations. 

We consider a simple random graph model for the generative model of random variables $\{w_{ij}\}_{i,j=1}^{n}$ given data vectors $\{\bs x_i\}_{i=1}^{n}$.
The conditional distribution of $w_{ij}$ is specified by a \emph{similarity function} $h(\bs x_i, \bs x_j)$ of the two data vectors.
Typically, Bernoulli distribution $P(w_{ij}=1 | \bs x_i, \bs x_j) = \sigma(h(\bs x_i,\bs x_j))$
with sigmoid function $\sigma(x):=(1+\exp(-x))^{-1}$,
and Poisson distribution $\text{Po}(\exp(h(\bs x_i,\bs x_j)))$ 
 are used to model the conditional probability.
These models are in fact specifying the conditional expectation $E(w_{ij}|\bs x_i, \bs x_j)$ by
$\sigma(h(\bs x_i,\bs x_j))$ and $\exp(h(\bs x_i,\bs x_j)) $, respectively, and they
correspond to logistic regression and Poisson regression in the context of generalized linear models.

We model the similarity function as
\begin{align}
h(\bs x_i,\bs x_j)
:=
g(f(\bs x_i),f(\bs x_j)), \label{eq:similarity}
\end{align}
where $f:\mathbb{R}^p \to \mathbb{R}^K$ is a continuous map and $g:\mathbb{R}^{K \times K} \to \mathbb{R}$ is a symmetric continuous function, which is defined later in Definition~\ref{def:kernel}. 
By using a neural network $\bs y = f_{\bs \psi}(\bs x)$ with parameter $\bs \psi$, 
we consider the model $h(\bs x_i,\bs x_j) = g( f_{\bs \psi}(\bs x_i),f_{\bs \psi}(\bs x_j)  )$,
which is called siamese network~\citep{bromley1994signature} in neural network literature.
The original form of siamese network uses the cosine similarity for $g$, but we can specify other types of similarity function.
By specifying the inner product $g(\bs y, \bs y') = \langle \bs y, \bs y' \rangle$,
the similarity function (\ref{eq:similarity}) becomes
\begin{align}
h(\bs x_i,\bs x_j)
=
\langle f_{\bs \psi}(\bs x_i),f_{\bs \psi}(\bs x_j) \rangle.
\label{eq:ips}
\end{align}
We call (\ref{eq:ips}) as Inner Product Similarity~(IPS) model. IPS commonly appears in a broad range of methods, such as DeepWalk~\citep{perozzi2014deepwalk}, LINE~\citep{tang2015line}, node2vec~\citep{grover2016node2vec}, 
Variational Graph AutoEncoder~\citep{kipf2016variational}, and GraphSAGE~\citep{hamilton2017inductive}. 
Multi-view extensions \citep{okuno2018probabilistic} are easily obtained by preparing different $f$ for each view and restricting loss terms in objective only to specific pairs; for example, the skip-gram model considers a  bipartite graph of two-views with the conditional distribution of contexts given a word.



\section{Previous study: PD similarities}
\label{sec:pd_kernels}



In order to prove the approximation capability of IPS given in eq.~(\ref{eq:ips}), \citet{okuno2018probabilistic} incorporates Universal Approximation Theorem of NN~\citep{funahashi1989approximate,cybenko1989approximation,yarotsky2016error,pmlr-v70-telgarsky17a} into Mercer's theorem~\cite{minh2006mercer}. 
To show the result in Theorem~\ref{theo:universal_approximate}, we first define a kernel and its positive-definiteness.

\begin{defi}
\normalfont
\label{def:kernel}
For some set $\mathcal{Y}$,
    a symmetric continuous function $g:\mathcal{Y}^2 \to \mathbb{R}$ is called a \textit{kernel} on $\mathcal{Y}^2$. 
\end{defi}

\begin{defi}
\normalfont
\label{def:pd}
    A kernel $g$ on $\mathcal{Y}^2$ is said to be \textit{Positive Definite~(PD)} if satisfying $\sum_{i=1}^{n}\sum_{j=1}^{n} c_i c_j g(\bs y_i,\bs y_j) \geq 0$ for arbitrary $c_1,c_2,\ldots,c_n \in \mathbb{R},\bs y_1,\bs y_2,\ldots,\bs y_n \in \mathcal{Y}$. 
\end{defi}

For instance, cosine similarity 
$
g(\bs y,\bs y'):=\langle \frac{\bs y}{\|\bs y\|_2}, \frac{\bs y'}{\|\bs y'\|_2} \rangle
$ 
is a PD kernel on $(\mathbb{R}^p \setminus \{\bs 0\})^2$. 
Its PD-ness immediately follows from $\sum_{i=1}^{n}\sum_{j=c}^{n} c_i c_j g(\bs y_i,\bs y_j)=\|\sum_{i=1}^{n} c_i \frac{\bs y_i}{\|\bs y_i\|_2} \|_2^2 \geq 0$ for arbitrary $\{c_i\}_{i=1}^{n} \subset \mathbb{R}$ and $\{\bs y_i\}_{i=1}^{n} \subset \mathcal{Y}$. 
Also polynomial kernel, Gaussian kernel, and Laplacian kernel are PD~\citep{berg1984harmonic}. 
By utilizing these kernels, we define a similarity of data vectors.

\begin{defi}
\normalfont 
\label{def:similarity}
	A function $h(\bs x,\bs x'):=g(f(\bs x),f(\bs x'))$ with a kernel $g:\mathcal{Y}^2 \to \mathbb{R}$ and a continuous map $f:\mathcal{X} \to \mathcal{Y}$ is called a \textit{similarity} on $\mathcal{X}^2$. 
\end{defi}

For a PD kernel $g$, the similarity $h$ is also a PD kernel on $\mathcal{X}^2$,
since  $\sum_{i=1}^{n}\sum_{j=1}^{n} c_i c_j h(\bs x_i,\bs x_j) = \sum_{i=1}^{n}\sum_{j=1}^{n} c_i c_j g(f(\bs x_i), f(\bs x_j)) \geq 0$.

Briefly speaking, a similarity $h$ is used for measuring how similar two data vectors are, while a kernel is used to compare feature vectors. 
Regarding PD similarities, the following Theorem~\ref{theo:universal_approximate} shows that IPS approximates any PD similarities arbitrary well if the number of hidden units and output dimension are sufficiently large. 

\begin{theo}[\citet{okuno2018probabilistic} \scalebox{0.9}{Theorem 5.1 $(D=1)$}] 
\label{theo:universal_approximate}
\normalfont
Let $f_*:[-M,M]^{p}\to \mathcal{Y}$ be a continuous function and $g_*:\mathcal{Y}^2 \to \mathbb{R}$ be a PD kernel for some closed set $\mathcal{Y} \subset \mathbb{R}^{K^*}$ and some $K^*, M>0$. 
$\sigma(\cdot)$ is ReLU or activation function which is non-constant, continuous, bounded, and monotonically-increasing.
Then, for arbitrary $\varepsilon>0$, 
by specifying sufficiently large $K \in \mathbb{N},T=T(K) \in \mathbb{N}$, 
there exist $\bs A \in \mathbb{R}^{K \times T},\bs B \in \mathbb{R}^{T \times p},\bs c \in \mathbb{R}^{T}$ such that
\begin{equation*}
\scalebox{0.9}{$
\bigg|
g_*\left(f_*(\x),f_*(\x')\right)
-
\big\langle f_{\bs \psi}(\x), f_{\bs \psi}(\x') \big\rangle
\bigg|
<\varepsilon
$}
\end{equation*}
for all $(\bs x,\bs x') \in [-M,M]^{2p}$, where $f_{\bs \psi}(\x)
=
\bs A \bs \sigma(\bs B \x + \bs c)$ is a two-layer neural network with $T$ hidden units and $K$ outputs 
and  $\bs\sigma(\x)$ is element-wise $\sigma(\cdot)$ function.
\end{theo}

The proof of Theorem~\ref{theo:universal_approximate} is based on the expansion
$
	g_*(f_*(\bs x),f_*(\bs x'))
	=
	\sum_{k=1}^{\infty} \lambda_k \tilde{\eta}_k(f_*(\bs x))\tilde{\eta}_k(f_*(\bs x')),
$ 
where $\tilde{\eta}_1,\tilde{\eta}_2,\ldots$ are eigenfunctions of $g_*$ whose eigenvalues are $\lambda_1 \geq \lambda_2 \geq \cdots \geq 0$. 
This expansion (a.k.a. Mercer's theorem) indicates with a vector-valued function $\bs \eta_K(\bs x):=(\lambda_1^{1/2}\tilde{\eta}_1(f_*(\bs x)),\ldots,\lambda_K^{1/2}\tilde{\eta}_K(f_*(\bs x)))$ that
$$
	\langle \bs \eta_K(\bs x), \bs \eta_K(\bs x') \rangle
	\to
	g_*(f_*(\bs x),f_*(\bs x')), 
	\:
	K \to \infty,
$$
for all $\bs x,\bs x'$. 
Considering a vector-valued NN $f_{\bs \psi}:\mathbb{R}^p \to \mathbb{R}^K$ that approximates $\bs \eta_K$, the IPS $\langle f_{\bs \psi}(\bs x),f_{\bs \psi}(\bs x') \rangle \approx \langle \bs \eta_K(\bs x),\bs \eta_K(\bs x') \rangle$ converges to $g_*(f_*(\bs x),f_*(\bs x'))$ as $K \to \infty$, thus proving the assertion. 

Unlike Mercer's theorem indicating only the existence of the feature map $\bs \eta_K$, 
Theorem~\ref{theo:universal_approximate} shows that it can be implemented as a neural network $f_{\bs \psi}$ so that the IPS $\langle f_{\bs \psi}(\bs x),f_{\bs \psi}(\bs x') \rangle$ eventually approximates the PD similarity $g_*(f_*(\bs x),f_*(\bs x'))$ arbitrary well.




\section{CPD similarities}
\label{sec:cpd_kernels}

Theorem~\ref{theo:universal_approximate} shows that IPS approximates any PD similarities arbitrary well. However, similarities in general are not always PD.
To deal with non-PD similarities, we consider a class of similarities based on Conditionally PD~(CPD) kernels~\citep{berg1984harmonic,scholkopf2001kernel} which include PD kernels as special cases. We then extend IPS to approximate CPD similarities.
Since we know that IPS has nice properties such as ``linguistic regularities'' of feature vector $\bs y$, our consideration will be focused on similarity models with kernels based on inner product.
In fact, according to the UAT applied to the whole $h(\bs x, \bs x')$,
a NN of the form $f_{\bs \psi}(\bs x, \bs x')$ approximates any similarities arbitrary well,
but we do \emph{not} attempt such an approach \emph{without} kernels based on inner product.

The remaining of this section is organized as follows. 
In Section~\ref{subsec:limit_NNFL}, we point out the fundamental limitation of IPS to approximate a non-PD similarity.  
In Section~\ref{subsec:cpd_kernels}, we define CPD kernels with some examples. 
In Section~\ref{subsec:proposed_models}, we propose a novel model named Shifted IPS~(SIPS), by extending the IPS model. 
In Section~\ref{subsec:interpretation}, we give interpretations of SIPS and its simpler variant C-SIPS.
In Section~\ref{subsec:rt_cpd}, we prove that SIPS approximates CPD similarities arbitrary well.


\subsection{Fundamental limitation of IPS}
\label{subsec:limit_NNFL}

Let us consider the negative squared distance~(NSD) $g(\bs y,\bs y')=-\|\bs y-\bs y'\|_2^2$ and the identity map $f(\bs x)=\bs x$. Then the similarity function
$$
h(\bs x,\bs x')
=
g(f(\bs x),f(\bs x'))
=
-\|\bs x-\bs x'\|_2^2 
$$
defined on $\mathbb{R}^{p} \times \mathbb{R}^p$ is not PD but CPD, which is defined later in Section~\ref{subsec:cpd_kernels}. 
Regarding the NSD similarity, Proposition~\ref{prop:dist_approximation_lower_bound} shows a strictly positive lower bound of approximation error for IPS. 

\begin{prop}
\label{prop:dist_approximation_lower_bound}
\normalfont
Let $\bs \Phi(p,K)$ denote the set of all continuous maps from $\mathbb{R}^p$ to $\mathbb{R}^K$. 
For all $M>0,p,K \in \mathbb{N}$, we have
\begin{align*}
&\inf_{\bs \phi \in \bs \Phi(p,K)}
\frac{1}{(2M)^{2p}}
\int_{[-M,M]^{p}}
\int_{[-M,M]^{p}} \\
&\hspace{3em}
\bigg|
	-\|\bs x-\bs x'\|_2^2
	-
	\langle
		\bs \phi(\bs x),\bs \phi(\bs x')
	\rangle
\bigg|
\diff \bs x
\diff \bs x'
\geq 
\frac{2pM^2}{3}.
\end{align*}
\end{prop}

The proof is in Supplement~\ref{subsec:proof_lower_bound}.

Since $\bs \Phi(p,K)$ represents the set of arbitrary continuous maps including neural networks, Proposition~\ref{prop:dist_approximation_lower_bound} indicates that IPS does not approximate NSD similarity arbitrary well, even if NN has a huge amount of hidden units with sufficiently large output dimension. 

\subsection{CPD kernels and similarities}
\label{subsec:cpd_kernels}
Here, we introduce similarities based on Conditionally PD~(CPD) kernels~\citep{berg1984harmonic,scholkopf2001kernel} to consider non-PD similarities which IPS does not approximate arbitrary well. 
We first define CPD kernels.

\begin{defi}
\normalfont
	A kernel $g$ on $\mathcal{Y}^2$ is called \textit{Conditionally PD~(CPD)} if satisfying $\sum_{i=1}^{n}\sum_{j=1}^{n} c_i c_j g(\bs y_i,\bs y_j) \geq 0$ for arbitrary $c_1,c_2,\ldots,c_n \in \mathbb{R},\bs y_1,\bs y_2,\ldots,\bs y_n \in \mathcal{Y}$ with the constraint $\sum_{i=1}^{n} c_i=0$. 
\end{defi}

The difference between the definitions of CPD and PD kernels is whether it imposes the constraint $\sum_{i=1}^{n} c_i=0$ or not.
According to these definitions, CPD kernels include PD kernels as special cases.
For a CPD kernel $g$, the similarity $h$ is also a CPD kernel on $\mathcal{X}^2$.

A simple example of CPD kernel is $g(\bs y,\bs y')=-\|\bs y-\bs y'\|_2^{\alpha}$ for $0<\alpha \leq 2$ defined on $\mathbb{R}^K \times \mathbb{R}^K$. 
Other examples are $-(\sin(y-y'))^2$ and
$-\bs 1_{(0,\infty)}(y+y')$ on $\mathbb{R} \times \mathbb{R}$. 
CPD-ness is a well-established concept with interesting properties \citep{berg1984harmonic}:
For any function $u(\cdot)$, $g(\bs y,\bs y')=u(\bs y)+ u(\bs y')$ is CPD. Constants are CPD.
The sum of two CPD kernels is also CPD. 
For CPD kernels $g$ with $g(\bs y,\bs y')\le0$, CPD-ness holds for  $-(-g)^{\alpha} \: (\alpha \in (0,1])$ and $-\log(1-g)$.

\begin{ex}[Poincar\'e distance] 
\label{ex:poincare_embedding}
\normalfont 
Let $B^K:=\{ \bs y \in \mathbb{R}^K \mid \|\bs y \|_2 <1\}$ be a $K$-dimensional open unit ball and define 
a distance between $\bs y,\bs y' \in B^K$ as
\begin{equation*}
\scalebox{0.9}{$\displaystyle 
d_{\text{Poincar\'e}}(\bs y,\bs y')
:=
\cosh^{-1}\left(
    1+2 \frac{\|\bs y-\bs y'\|_2^2}{(1-\|\bs y\|_2^2)(1-\|\bs y'\|_2^2)}
\right),$}
\end{equation*}
where $\cosh^{-1}(z)=\log(z+\sqrt{z+1}\sqrt{z-1})$. 
Considering the setting of Section~\ref{sec:background} with 1-hot data vectors, Poincar\'e embedding~\citep{nickel2017poincare} learns parameters $\bs y_i$, $i=1,\ldots,n$,
by fitting $\sigma(-d_{\text{Poincar\'e}}(\bs y_i, \bs y_j))$ to the observed $w_{ij} \in \{0,1\}$. 

\end{ex}

Interestingly, negative Poincar\'e distance is proved to be CPD in \citet[Corollary 7.4]{faraut1974distances}. 

\begin{prop}
\label{prop:poincare_cpd}
\normalfont
$-d_{\text{Poincar\'e}}$ is CPD on $B^K \times B^K$. 
\end{prop}

It is strictly CPD in the sense that $-d_{\text{Poincar\'e}}$ is not PD.
A counter-example of PD-ness is, for example, $n=2,K=2,c_1=c_2=1,\bs y_1=(1/2,1/2),\bs y_2=(0,0) \in B^2$.

Another interesting example of CPD kernels is negative Wasserstein distance. 

\begin{ex}[Wasserstein distance] 
\label{ex:wasserstein_embedding}
\normalfont 
For $q \in (0,\infty)$, 
let $\bs Z$ be a metric space endowed with a metric $d_Z$, which we call as ``ground distance''. 
Let $\mathcal{Y}$ be the space of all measures $\mu$ on $\bs Z$  satisfying $\int_{\bs Z} d_Z(\bs z,\bs z_0) \diff \mu(\bs z) <\infty$ for all $\bs z_0 \in \bs Z$. 
The $q$-Wasserstein distance between $\bs y,\bs y'$ is defined as
\begin{equation*}
\scalebox{0.9}{$ \displaystyle
d_{W}^{(q)}(\bs y,\bs y')
:=
\left(
	\inf_{\pi \in \Pi(\bs y,\bs y')}
	\iint_{\bs Z \times \bs Z}
	d_Z(\bs z,\bs z')^q
	\diff \pi(\bs z,\bs z')
\right)^{1/q}$}.
\end{equation*}
Here, $\Pi(\bs y,\bs y')$ is the set of joint probability measures on $\bs Z \times \bs Z$ having marginals $\bs y,\bs y'$. 
Wasserstein distance is used for a broad range of methods, such as Generative Adversarial Networks~\citep{arjovsky2017wasserstein} and AutoEncoder~\citep{tolstikhin2018wasserstein}. 

%
\end{ex}

With some assumptions, negative Wasserstein distance is proved to be CPD. 

\begin{prop}
\label{prop:wasserstein_cpd}
\normalfont
$-d_W^{(1)}$ is CPD on $\mathcal{Y}^2$ if $-d_Z$ is CPD on $\bs Z^2$. 
$-d_W^{(2)}$ is CPD on $\mathcal{Y}^2$ if $\bs Z$ is a subset of $\mathbb{R}$.
\end{prop}

$-d_W^{(1)}$ is known as the negative earth mover's distance, and its CPD-ness is discussed in \citet{gardner2017definiteness}. 
The CPD-ness of $-d_W^{(2)}$ is shown in \citet[Corollary 1]{kolouri2016sliced}. 


Therefore negative Poincar\'e distance and negative Wasserstein distance are CPD kernels. 
In the following section, we propose a novel model that approximates any CPD similarities arbitrary well. 



\subsection{Proposed models}
\label{subsec:proposed_models}
For extending IPS model given in eq.~(\ref{eq:ips}), we propose a novel model
\begin{align}
h(\bs x_i, \bs x_j) = 
\langle f_{\bs \psi}(\bs x_i),f_{\bs \psi}(\bs x_j) \rangle + u_{\bs \xi}(\bs x_i) + u_{\bs \xi}(\bs x_j),
\label{eq:sips}
\end{align}
where $f_{\bs \psi}:\mathbb{R}^p \to \mathbb{R}^K$ and $u_{\bs \xi}:\mathbb{R}^p \to \mathbb{R}$ are neural networks whose parameter matrices are $\bs \psi$ and $\bs \xi$, respectively. 
We call (\ref{eq:sips}) as Shifted IPS~(SIPS) model,
because the inner product $\langle f_{\bs \psi}(\bs x_i),f_{\bs \psi}(\bs x_j) \rangle$ is
shifted by the offset $u_{\bs \xi}(\bs x_i) + u_{\bs \xi}(\bs x_j)$.
Later, we show in Theorem~\ref{theo:universal_cpd} that SIPS approximates any CPD kernels arbitrary well.

We also consider a special case of SIPS. 
By assuming $u_{\bs \xi}(\bs x)=-\gamma/2$ for all $\bs x$, SIPS reduces to 
\begin{align}
h(\bs x_i, \bs x_j) = 
\langle f_{\bs \psi}(\bs x_i),f_{\bs \psi}(\bs x_j) \rangle -\gamma, \label{eq:csips}
\end{align}
where $\gamma \geq 0$ is a parameter to be estimated. 
We call (\ref{eq:csips}) as Constantly-Shifted IPS~(C-SIPS) model.

If we have no attributes, we use 1-hot vectors for $\bs x_i$ in $\mathbb{R}^n$ instead,
and $f_{\bs \psi}(\bs x_i) = {\bs y}_i \in \mathbb{R}^K$, $u_{\bs \xi}(\bs x_i) = u_i \in \mathbb{R}$ are
model parameters. Then SIPS reduces to
the matrix decomposition model with biases
\begin{equation} \label{eq:matrix-factorization-with-bias}
h(\bs x_i, \bs x_j) = \langle {\bs y}_i, {\bs y}_j \rangle + u_i + u_j.
\end{equation}
This model is widely used for recommender systems \citep{koren2009matrix} and word vectors \citep{pennington2014glove}, and SIPS is considered as its generalization.

\subsection{Interpretation of SIPS and C-SIPS}
\label{subsec:interpretation}
Here we illustrate the interpretation of the proposed models by returning back to the setting in Section~\ref{sec:background}. 
We consider a simple generative model of independent Poisson distribution with mean parameter $E(w_{ij}) = \exp(h(\bs x_i,\bs x_j))$.
Then SIPS gives a generative model
\begin{align}
w_{ij}
\overset{\text{indep.}}{\sim}
\text{Po}\Bigl(
	\beta(\bs x_i)
	\beta(\bs x_j)	
	\exp(\langle f_{\bs \psi}(\bs x_i),f_{\bs \psi}(\bs x_j) \rangle)\Bigr), \label{eq:sips_weighted_model}
\end{align}
where $\beta(\bs x):=\exp(u_{\bs \psi}(\bs x))>0$. 
Since $\beta(\bs x)$ can be regarded as the ``importance weight'' of data vector $\bs x$, 
SIPS naturally incorporates the weight function $\beta(\bs x)$ to probabilistic models used in a broad range of existing methods.
Similarly, C-SIPS gives a generative model
\begin{align}
w_{ij}
\overset{\text{indep.}}{\sim}
\text{Po}\Bigl(
	\alpha
	\exp(\langle f_{\bs \psi}(\bs x_i),f_{\bs \psi}(\bs x_j) \rangle) \Bigr), \label{eq:csips_weighted_model}
\end{align}
where $\alpha:=\exp(-\gamma)>0$ regulates the sparseness of $\{w_{ij}\}$.
The generative model (\ref{eq:csips_weighted_model}) is already proposed as 1-view PMvGE~\citep{okuno2018probabilistic}.

It was shown in Supplement~C of \citet{okuno2018probabilistic} that PMvGE (based on C-SIPS) approximates CDMCA
when $w_{ij}$ is replaced by $\delta_{ij}$ in the constraint (8) therein,
and this result can be extended so that PMvGE with SIPS approximates the original CDMCA using $w_{ij}$ in the constraint.

\subsection{Representation theorems}
\label{subsec:rt_cpd}

Theorem~\ref{theo:universal_cpd} below shows that SIPS given in eq.~(\ref{eq:sips})
approximates any CPD similarities arbitrary well and thus it overcomes the fundamental limitation of IPS.  
Theorem~\ref{theo:universal_cpd2} proves that C-SIPS given in eq.~(\ref{eq:csips})
also approximates CPD similarities in a weaker sense.

\begin{theo}[Representation theorem for SIPS]
\normalfont
\label{theo:universal_cpd}
Let $f_*:[-M,M]^{p}\to \mathcal{Y}$ be a continuous function and $g_*:\mathcal{Y}^2 \to \mathbb{R}$ be a CPD kernel for some closed set $\mathcal{Y} \subset \mathbb{R}^{K^*}$ and some $K^*, M>0$. 
$\sigma(\cdot)$ is ReLU or activation function which is non-constant, continuous, bounded, and monotonically-increasing.
Then, for arbitrary $\varepsilon>0$, 
by specifying sufficiently large $K \in \mathbb{N},T=T(K),T' \in \mathbb{N}$, 
there exist $\bs A \in \mathbb{R}^{K \times T},\bs B \in \mathbb{R}^{T \times p},\bs c \in \mathbb{R}^{T},
\bs e \in \mathbb{R}^{T'}, \bs F \in \mathbb{R}^{T' \times p}, \bs o \in \mathbb{R}^{T'}$ such that
\begin{align*}
&\scalebox{0.9}{$
\bigg|
g_*\left(f_*(\x),f_*(\x')\right)$} \nonumber \\
&\hspace{3em}
\scalebox{0.9}{$
-
\left(\big\langle f_{\bs \psi}(\x), f_{\bs \psi}(\x') \big\rangle
+
u_{\bs \xi}(\x)
+
u_{\bs \xi}(\x')
\right)
\bigg|
<\varepsilon$}
\end{align*}
for all $(\bs x,\bs x') \in [-M,M]^{2p}$, where $f_{\bs \psi}(\x)
=
\bs A \bs \sigma(\bs B\x + \bs c) \in \mathbb{R}^{K}$ and
$u_{\bs \xi}(\x)
=
\langle \bs e, \bs \sigma(\bs F \x+\bs o) \rangle \in \mathbb{R}$ are two-layer neural networks with $T$ and $T'$ hidden units, respectively, 
and  $\bs\sigma(\x)$ is element-wise $\sigma(\cdot)$ function.
\end{theo}

The proof stands on Lemma 2.1 in \citet{berg1984harmonic}, which indicates the equivalence between 
CPD-ness of $g_*(\bs y,\bs y')$ and 
PD-ness of $g_0(\bs y,\bs y'):=g_*(\bs y,\bs y')-g_*(\bs y,\bs y_0)-g_*(\bs y_0,\bs y')+g_*(\bs y_0,\bs y_0)$ with fixed $\bs y_0 \in \mathcal{Y}$. 
Here, we consider a NN $f_{\bs \psi}(\bs x)$ such that $\langle f_{\bs \psi}(\bs x),f_{\bs \psi}(\bs x') \rangle$ approximates $g_0(f_*(\bs x),f_*(\bs x'))$. Such a NN $f_{\bs \psi}$ is guaranteed to exist, due to Theorem~\ref{theo:universal_approximate} and the PD-ness of $g_0$. 
By considering another NN $u_{\bs \xi}(\bs x)$ that approximates $g_*(f_*(\bs x),\bs y_0)-\frac{1}{2}g_*(\bs y_0,\bs y_0)$, we have 
$g_*(f_*(\bs x),f_*(\bs x')) =
g_0(f_*(\bs x),f_*(\bs x'))+g_*(f_*(\bs x),\bs y_0)+g_*(\bs y_0,f_*(\bs x'))-g_*(\bs y_0,\bs y_0)
\approx
\langle f_{\bs \psi}(\bs x),f_{\bs \psi}(\bs x') \rangle
+
u_{\bs \xi}(\bs x)
+
u_{\bs \xi}(\bs x')$, thus proving the assertion. 
The detailed proof is in Supplement~\ref{sec:proof_universal_cpd}. 
%
%

\begin{theo}[Representation theorem for C-SIPS]
\normalfont
\label{theo:universal_cpd2}
Symbols and assumptions are the same as those of Theorem~\ref{theo:universal_cpd}
but $\mathcal{Y}$ is a compact set.
For arbitrary $\varepsilon>0$, 
by specifying sufficiently large $K \in \mathbb{N}$, $T=T(K) \in \mathbb{N}$, $r>0$, 
there exist $\bs A \in \mathbb{R}^{K \times T}$, $\bs B \in \mathbb{R}^{T \times p}$, $\bs c \in \mathbb{R}^{T}$, $\gamma=O(r^2)$ such that
\begin{align*}
&
\scalebox{0.9}{$
\bigg|
g_*\left(f_*(\x),f_*(\x')\right)$} \nonumber \\
&\hspace{3em}
\scalebox{0.9}{$-
\left(\big\langle f_{\bs \psi}(\x), f_{\bs \psi}(\x') \big\rangle
-\gamma\right)
\bigg|
<\varepsilon+O(r^{-2})$}
\end{align*}
for all $(\bs x,\bs x') \in [-M,M]^{2p}$, where $f_{\bs \psi}(\x)
=
\bs A \bs \sigma(\bs B\x + \bs c) \in \mathbb{R}^{K}$ is a two-layer neural network with $T$ hidden units.
\end{theo}
The proof is in Supplement~\ref{sec:proof_universal_cpd2}. 

For reducing the approximation error of order $O(r^{-2})$ of C-SIPS in Theorem~\ref{theo:universal_cpd2}, we will have a large $r$.
Then large $\gamma=O(r^2)$ value leads to unstable computation of NN as shown in Section~\ref{sec:experiment}. 
Conversely, a small $r$ increases the upper bound of the approximation error. 
Thus, if available, we prefer SIPS in terms of both computational stability and small approximation error. 

\section{Numerical experiment}
\label{sec:experiment}

In this section, we conduct a numerical experiment on synthetic data to compare existing model~(IPS), our novel model~(SIPS), and its simper variant~(C-SIPS).
The experiment settings are explained in Section~\ref{subsec:exp_setting}, and the results are shown in Section~\ref{subsec:exp_result}. 

\subsection{Settings}
\label{subsec:exp_setting}

\textbf{Kernels:}
Three types of kernels are considered as $g_*(\bs y,\bs y')$ for generating simulation data:
(i) cosine similarity $\langle \frac{\bs y}{\|\bs y\|_2},\frac{\bs y'}{\|\bs y'\|_2} \rangle$, 
(ii) negative squared distance $-\|\bs y-\bs y'\|_2^2$, 
(iii) negative Poincar\'e distance $-d_{\text{Poincar\'e}}(\bs y,\bs y')$ defined in
Example~\ref{ex:poincare_embedding}.
These kernels are PD, CPD, and CPD, in this order.

\textbf{Synthetic data:} 
For kernels (i) and (ii), 
data vectors $\{\bs x_i\} \subset \mathbb{R}^p$ are generated independently by uniform distribution over $[-2,2]^p$ with $p=5$. 
Feature vectors of dimensions $K^*=4$ are computed by a continuous map $f_*(\bs x):=(x_1,\cos x_2,\exp(-x_3),\sin(x_4-x_5)) \in \mathbb{R}^4$, and similarity values are given as
\begin{align}
h_{ij}^* := g_*(f_*(\bs x_i),f_*(\bs x_j))
\label{eq:true_similarity}
\end{align}
for all $1 \leq i,j \leq n$. 
For kernel (iii), data vectors $\{\bs x_i\} \subset \mathbb{R}^{p}$ are generated by $\displaystyle \bs x_i:=r_i \tilde{\bs x}_i/\|\tilde{\bs x}_i\|_2 $ with $\tilde{\bs x}_i \overset{\text{i.i.d.}}{\sim} N_p(\bs 0,\bs I),r_i \overset{\text{i.i.d.}}{\sim} B(5,1)$, so that $\|\bs x_i\|_2<1$. $B(\alpha,\beta)$ is the beta distribution with parameters $\alpha,\beta>0$, and $\mathcal{N}_p(\bs 0,\bs I)$ represents the $p$-variate standard normal distribution. 
Feature vectors of dimensions $K^*=p=5$ are computed by the identity map $f_*(\bs x)=\bs x$,
and similarity values $\{h_{ij}^*\}$ are given as (\ref{eq:true_similarity}). 
In order to simplify the experiment just for illustrating the differences of similarity models,
we do not generate $\{ w_{ij} \}$ and treat $\{h_{ij}^*\}$ as observed samples.
For each setting (i)-(iii), we generate $n=1{,}000$ training samples and $n=3{,}000$ test samples.

\textbf{NN architecture:} Three models are considered for similarity function:
(i) IPS~(existing) defined in eq.~(\ref{eq:ips}), 
(ii) SIPS~(proposed) defined in eq.~(\ref{eq:sips}), 
(iii) C-SIPS~(the simpler variant of SIPS) defined in eq.~(\ref{eq:csips}). 
For each model, $f_{\bs \psi}:\mathbb{R}^p \to \mathbb{R}^K,u_{\bs \xi}:\mathbb{R}^p \to \mathbb{R}$ are two-layer NNs with $T$ hidden units and ReLU activations. We denote $T$ as ``$\sharp$units''.

\textbf{Training:} 
For training the three similarity models, we minimize the mean squared error between $h_{ij}^*$
and $h(\bs x_i, \bs x_j)$.
The loss functions are $\ell^2$-regularized with a coefficient $0.01$, 
and they are minimized by $10{,}000$ iterations of full-batch gradient descent. 
The learning rate is starting from $0.001$ and attenuated by $1/10$ for every $100$ iterations.

\textbf{Evaluation:} Neural networks are trained with $1{,}000$ samples, and they are evaluated by the Mean Squared Prediction Error~(MSPE) with respect to $3{,}000$ test samples. 
We compute the average and the standard deviation of 5 runs for each setting.

\subsection{Result}
\label{subsec:exp_result}

In the following plots, black, blue, and red lines represent the MSPE of IPS~(Existing), SIPS~(Proposed), and its simpler variant C-SIPS, respectively. 

Table~\ref{tab:exp_cosine} shows the MSPE for the cosine similarity. 
Error bar shows the standard deviation (=$1\sigma$). 
In accordance with the theory, all of IPS, SIPS, and C-SIPS show the good approximation performance since cosine similarity is PD. 
Interestingly, output dimension $K=3$ is sufficient to approximate the function $g_*(f_*(\bs x),f_*(\bs x'))$ regardless of the number of hidden units.


Table~\ref{tab:exp_nsd} shows the MSPE for the negative squared distance~(NSD). 
In accordance with the theory, NSD is well approximated by SIPS but not by IPS due to its CPD-ness. 
The approximation error of SIPS with $m=1{,}000$ becomes almost zero at $K=4$ as expected from $K^*=4$.
In theory, C-SIPS can approximate CPD kernels, but it does not perform well in this setting.
Since C-SIPS requires the parameter $\gamma$ to be very large for minimizing the approximation error, 
its computation becomes unstable in some cases. 

Table~\ref{tab:exp_poincare} shows the MSPE for $-d_{\text{Poincar\'e}}$. 
In accordance with the theory, the generated similarity values are well approximated by both SIPS and C-SIPS, but not by IPS due to the CPD-ness.
The approximation error of SIPS and C-SIPS with $m\ge100$ becomes almost zero at $K=5$ as expected from $K^*=5$.
However, the standard deviation of MSPE for C-SIPS is large with $T=10$.

\begin{table}[htbp]
\caption{ Cosine similarity: $\langle \frac{\bs y}{\|\bs y\|_2},\frac{\bs y'}{\|\bs y'\|_2} \rangle$
}
\label{tab:exp_cosine}
\vspace{1ex}
\centering
\subfigure[$\sharp$units$=10$]{
	\includegraphics[width=2.5cm]{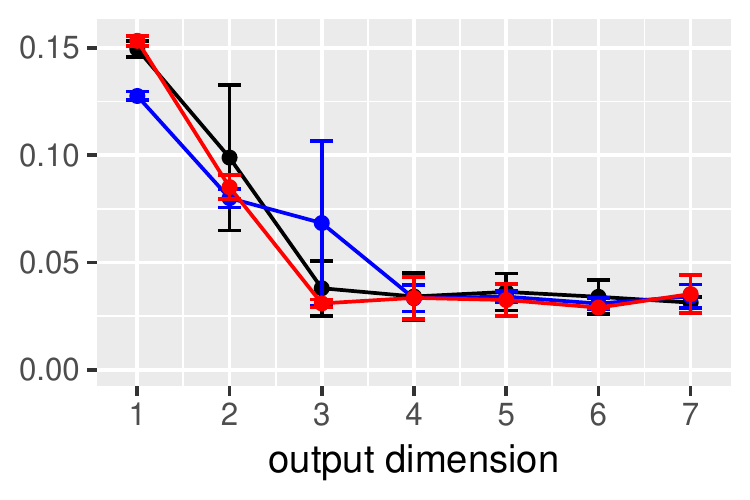}
	\label{fig:cosine10}
}
\subfigure[$\sharp$units$=100$]{
	\includegraphics[width=2.5cm]{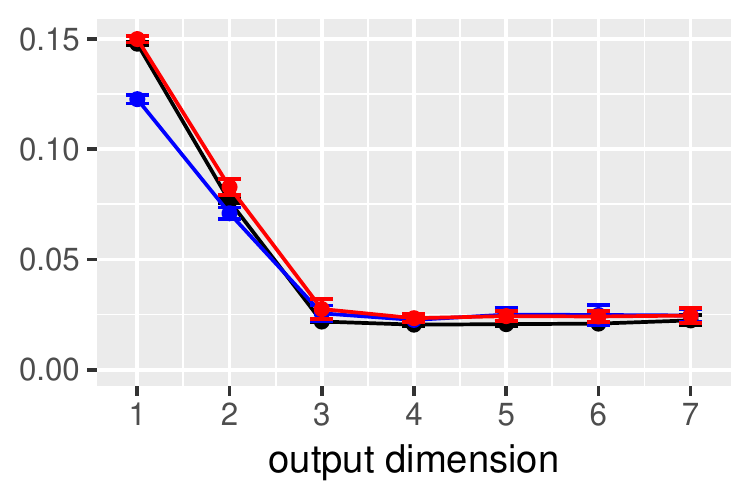}
	\label{fig:cosine100}
}
\subfigure[$\sharp$units$=1000$]{
	\includegraphics[width=2.5cm]{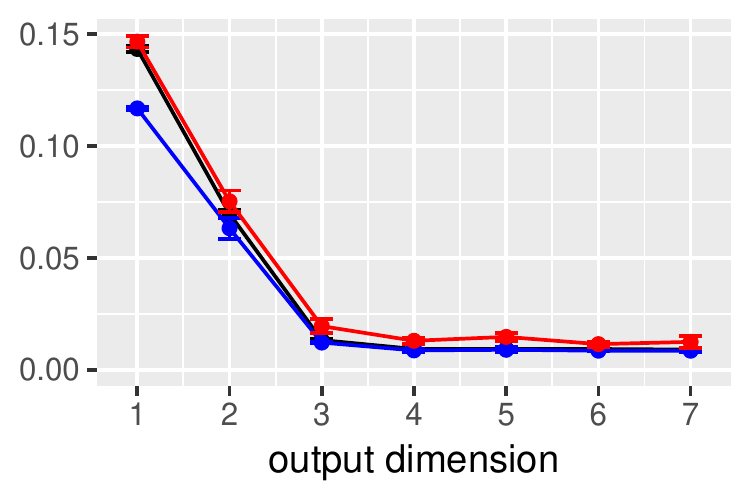}
	\label{fig:cosine1000}
}
\vspace*{-4ex}
\end{table}

\begin{table}[htbp]
\caption{Negative squared distance: $-\|\bs y-\bs y'\|_2^2$
}
\label{tab:exp_nsd}
\vspace{1ex}
\centering
\subfigure[$\sharp$units$=10$]{
	\includegraphics[width=2.5cm]{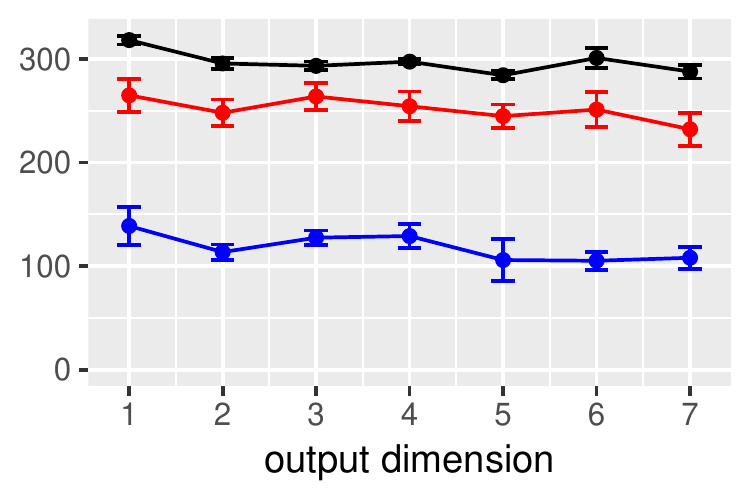}
	\label{fig:nsd10}
}
\subfigure[$\sharp$units$=100$]{
	\includegraphics[width=2.5cm]{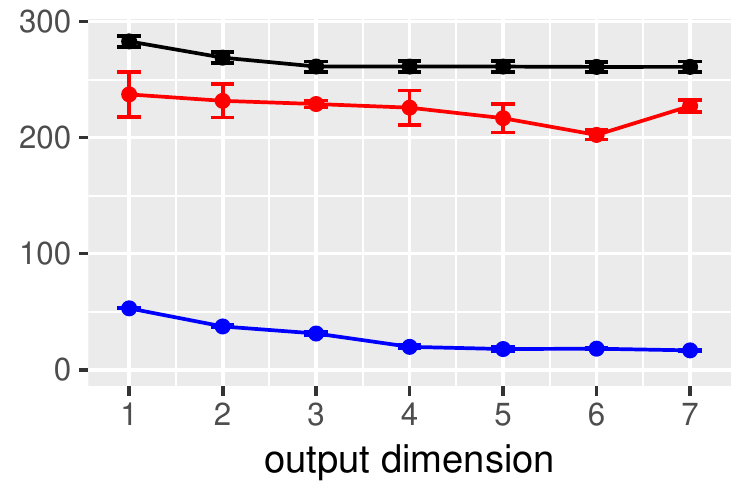}
	\label{fig:nsd100}
}
\subfigure[$\sharp$units$=1000$]{
	\includegraphics[width=2.5cm]{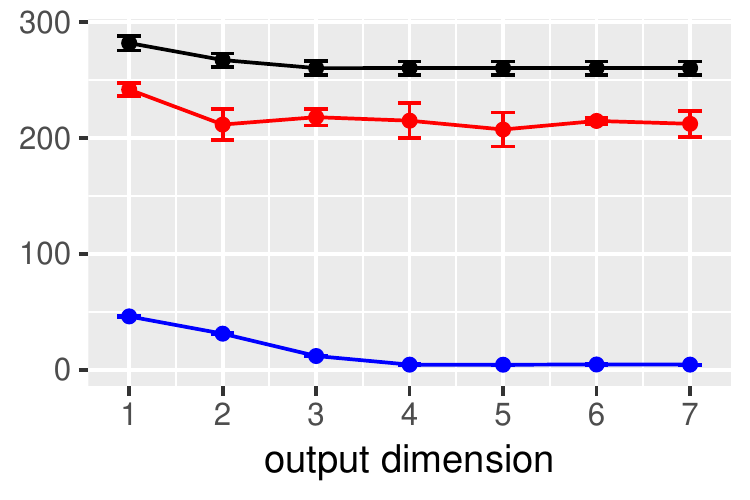}
	\label{fig:nsd1000}
}
\vspace*{-4ex}
\end{table}

\begin{table}[htbp]
\caption{
Negative Poincar\'e distance: $-d_{\text{Poincar\'e}}(\bs y,\bs y')$
}
\label{tab:exp_poincare}
\vspace{1ex}
\centering
\subfigure[$\sharp$units$=10$]{
	\includegraphics[width=2.5cm]{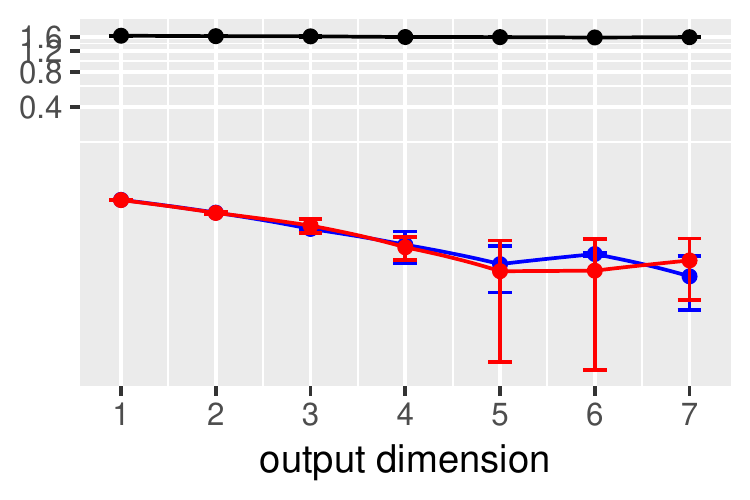}
	\label{fig:poincare10}
}
\subfigure[$\sharp$units$=100$]{
	\includegraphics[width=2.5cm]{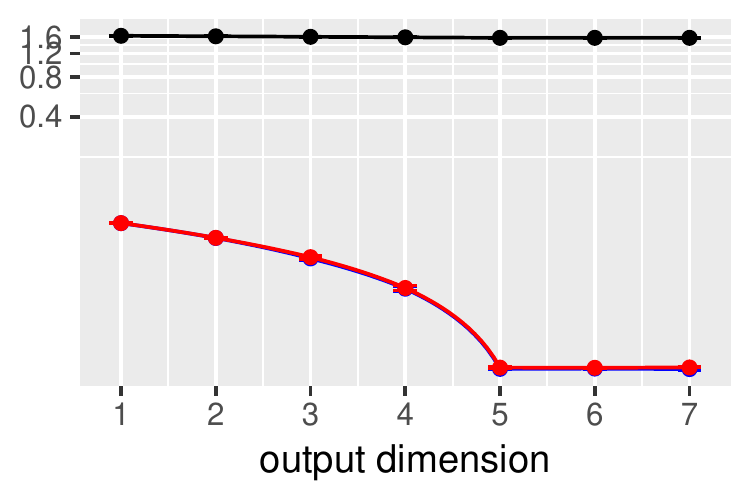}
	\label{fig:poincare100}
}
\subfigure[$\sharp$units$=1000$]{
	\includegraphics[width=2.5cm]{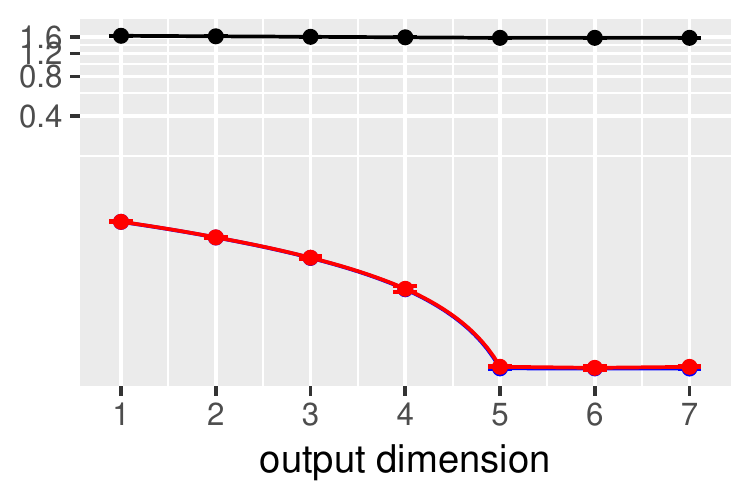}
	\label{fig:poincare1000}
}
\end{table}

\section{Conclusion}
\label{sec:conclusion}
In this paper, we have considered the representation power of inner-product similarity~(IPS), Shifted IPS~(SIPS), and Constantly-SIPS~(C-SIPS). 
We have first pointed out the fundamental limitation of IPS to approximate non-PD similarities. 
To deal with such non-PD similarities, we have considered similarities based on CPD kernels, which include PD kernels as special cases, and we have proposed a novel model named SIPS by extending IPS. 
Then we proved that SIPS is capable of approximating any CPD similarities arbitrary well. 
Since negative Poincar\'e distance and negative Wasserstein distance are CPD, the similarities based on these distances can be approximated by SIPS. 
We have performed numerical experiments to show the superiority of SIPS over IPS. 


\appendix

\section{Further extension beyond CPD: General similarities}
\label{sec:arbitrary_kernels}
CPD includes a broad range of kernels, yet there exists a variety of non-CPD kernels. 
One example is Epanechnikov kernel $g(\bs y,\bs y'):=(1-\|\bs y-\bs y'\|_2^2)\bs 1(\|\bs y-\bs y'\|_2 \leq 1)$ defined on $\mathbb{R}^p \times \mathbb{R}^p$. 
To approximate similarities based on such non-CPD kernels, we propose an inner-product based model that has a high representation capability.
Although this model is not always easy to compute due to the excessive degrees of freedom, the model is, in theory, shown to be capable of approximating more general kernels that are considered in \citet{ong2004learning}.

\subsection{Proposed model}

Let us consider a similarity $h(\bs x,\bs x')=g_*(f_*(\bs x),f_*(\bs x'))$ with any kernel $g_*:\mathbb{R}^{2K^*} \to \mathbb{R}$ and a continuous map $f_*:\mathbb{R}^p \to \mathbb{R}^{K^*}$.
To approximate it, we consider a similarity model  
\begin{align}
h(\bs x_i, \bs x_j) = 
\langle f_{\bs \psi}(\bs x_i),f_{\bs \psi}(\bs x_j) \rangle - \langle r_{\bs \zeta}(\bs x_i),r_{\bs \zeta}(\bs x_j) \rangle,
\label{eq:general_ips}
\end{align}
where $f_{\bs \psi}:\mathbb{R}^{p} \to \mathbb{R}^{K_+}$ and $r_{\bs \zeta}:\mathbb{R}^p \to \mathbb{R}^{K_-}$ are neural networks whose parameters are $\bs \psi$ and $\bs \zeta$, respectively. 
Since the kernel $g(\bs y, \bs y') = \langle \bs y_+,\bs y'_+ \rangle - \langle \bs y_-,\bs y'_- \rangle$ with respect to $\bs y = (\bs y_+,\bs y_-) \in \mathbb{R}^{K_+ + K_-}$ is known as the inner product in Minkowski space~\citep{naber2012geometry}, we call (\ref{eq:general_ips}) as Minkowski IPS~(MIPS) model.

By replacing $f_{\bs \psi}(\bs x)$ and $r_{\bs \zeta}(\bs x)$ with 
$(f_{\bs \psi}(\bs x)^{\top},u_{\bs \xi}(\bs x),1)^{\top}$ and $u_{\bs \xi}(\bs x)-1 \in \mathbb{R}$, respectively, MIPS reduces to SIPS defined in eq.~(\ref{eq:sips}), meaning that MIPS includes SIPS as a special case. 
Therefore, MIPS approximates any CPD similarities arbitrary well. 
Further, we prove that MIPS approximates more general similarities arbitrary well.

\subsection{Representation theorem}

\begin{theo}[Representation theorem for MIPS]
\normalfont
\label{theo:universal_general}
Symbols and assumptions are the same as those of Theorem~\ref{theo:universal_cpd} but $g_*$ is a general kernel, which is only required to be dominated by some PD kernels $g$ (i.e., $g-g_*$ is PD). 
For arbitrary $\varepsilon>0$, 
by specifying sufficiently large $K_+, K_- \in \mathbb{N},T_+=T_+(K_+),T_-=T_-(K_-) \in \mathbb{N}$, 
there exist $\bs A \in \mathbb{R}^{K_+ \times T_+},\bs B \in \mathbb{R}^{T_+ \times p},\bs c \in \mathbb{R}^{T_+},
\bs E \in \mathbb{R}^{K_- \times T_-}, \bs F \in \mathbb{R}^{T_- \times p}, \bs o \in \mathbb{R}^{T_-}$ such that
\begin{align*}
&\scalebox{0.9}{$
\bigg|
g_*\left(f_*(\x),f_*(\x')\right)$} \nonumber \\
&\hspace{3em}
\scalebox{0.9}{$
-
\left(\big\langle f_{\bs \psi}(\x), f_{\bs \psi}(\x') \big\rangle
-\big \langle r_{\bs \zeta}(\x),r_{\bs \zeta}(\x') \big \rangle
\right)
\bigg|
<\varepsilon$}
\end{align*}
for all $(\bs x,\bs x') \in [-M,M]^{2p}$, where $f_{\bs \psi}(\x)
=
\bs A \bs \sigma(\bs B\x + \bs c) \in \mathbb{R}^{K_+}$ and
$r_{\bs \zeta}(x)
=
\bs E \bs \sigma(\bs F\x + \bs o) \in \mathbb{R}^{K_-}$ are two-layer neural networks with $T_+$ and $T_-$ hidden units, respectively, 
and  $\bs\sigma(\x)$ is element-wise $\sigma(\cdot)$ function.
\end{theo}

In theorem~\ref{theo:universal_general}, the kernel $g_*$ is only required to be dominated by some PD kernels, thus $g_*$ is not limited to CPD. 
Our proof for Theorem~\ref{theo:universal_general} is based on \citet[Proposition~7]{ong2004learning}.
This proposition indicates that the kernel $g_*$ dominated by some PD kernels is decomposed as the difference of two PD kernels $g_+,g_-$ by considering Krein space consisting of two Hilbert spaces.
Therefore, we have $g_*(f_*(\bs x),f_*(\bs x')) = g_+(f_*(\bs x),f_*(\bs x')) - g_-(f_*(\bs x),f_*(\bs x'))$. 
Because of the PD-ness of $g_+$ and $g_-$, 
Theorem~\ref{theo:universal_approximate} guarantees the existence of NNs $f_{\bs \psi},r_{\bs \zeta}$ such that 
$\langle f_{\bs \psi}(\bs x),f_{\bs \psi}(\bs x') \rangle$ and 
$\langle r_{\bs \zeta}(\bs x),r_{\bs \zeta}(\bs x') \rangle$, respectively, approximate 
$g_+(f_*(\bs x),f_*(\bs x'))$ and $g_-(f_*(\bs x),f_*(\bs x'))$ arbitrary well. Thus proving the theorem. 
This idea for the proof is also interpreted as a generalized Mercer's theorem for Krein space (there is a similar attempt in \citet{chen2008generalized}) by applying Mercer's theorem to the two Hilbert spaces of \citet[Proposition~7]{ong2004learning}.

\subsection{Deep Gaussian embedding}
To show another example of non-CPD kernels, Deep Gaussian embedding~\citep{bojchevski2018deep} is reviewed below.

\begin{ex}[Deep Gaussian embedding] 
\label{ex:deep_gaussian_embedding}
\normalfont 
Let $\mathcal{Y}$ be a set of distributions over a set $\bs Z \subset \mathbb{R}^q$. 
Kullback-Leibler divergence~\citep{kullback1951information} between two distributions $\bs y,\bs y' \in \mathcal{Y}$ is defined by 
$$
    d_{\text{KL}}(\bs y , \bs y')
    :=
    \int_{\bs Z} y(\bs z) \log \frac{y(\bs z)}{y'(\bs z)} \diff \bs z, 
$$
where $y(\bs z)$ is the probability density function corresponding to the distribution $\bs y \in \mathcal{Y}$.

With the same setting in Section~\ref{sec:background}, Deep Gaussian embedding~\citep{bojchevski2018deep}, which incorporates neural networks into Gaussian embedding~\citep{vilnis2014word}, learns two neural networks $\bs \mu:\mathbb{R}^p \to \mathbb{R}^q,\bs \Sigma:\mathbb{R}^p \to \mathbb{R}^{q \times q}_+$ so that 
the function $\sigma(-d_{\text{KL}}(
\mathcal{N}_q(\bs \mu(\bs x_i),\bs \Sigma(\bs x_i)),
\mathcal{N}_q(\bs \mu(\bs x_j),\bs \Sigma(\bs x_j)) ))$ 
approximates $E(w_{ij} | \bs x_i, \bs x_j)$. 
$\mathbb{R}_+^{q \times q}$ is a set of all $q \times q$ positive definite matrices and $\mathcal{N}_q(\bs \mu,\bs \Sigma)$ represents the $q$-variate normal distribution with mean $\bs \mu$ and variance-covariance matrix $\bs \Sigma$.

Unlike typical graph embedding methods, deep Gaussian embedding maps data vectors to distributions as
$$
    \mathbb{R}^p \ni \bs x \mapsto \bs y:=\mathcal{N}_q(\bs \mu(\bs x),\bs \Sigma(\bs x)) \in \mathcal{Y},
$$
where $\bs y$ is also interpreted as a vector of dimension $K=q+q(q+1)/2$ by considering the number of parameters in $\bs \mu$ and $\bs \Sigma$.
Our concern is to clarify if $d_{\text{KL}}$ is CPD. However, in the first place, $d_{\text{KL}}$ is not a kernel since it is not symmetric. 
In order to make it symmetric, Kullback-Leibler divergence may be replaced with Jeffrey's divergence~\citep{kullback1951information}
$$
d_{\text{Jeff}}(\bs y,\bs y')
:=
d_{\text{KL}}(\bs y,\bs y')
+
d_{\text{KL}}(\bs y',\bs y).
$$
\end{ex}

Although $-d_{\text{Jeff}}$ is a kernel, it is not CPD as shown in Proposition~\ref{prop:gaussian_not_cpd}. 

\begin{prop}
\label{prop:gaussian_not_cpd}
\normalfont
$-d_{\text{Jeff}}$ is not CPD on $\tilde{\mathcal{P}}_K^2$, where $\tilde{\mathcal{P}}_K$ represents the set of all $K$-variate normal distributions. 
\end{prop}

A counterexample of CPD-ness is, $n=3,q=2,
c_1=-2/5,c_2=-3/5,c_3=1, 
\bs y_i=\mathcal{N}_2(\bs \mu_i,\bs \Sigma_i) \in \mathcal{Y} \: (i=1,2,3),
\bs \mu_1=(2,1)^{\top},
\bs \mu_2=(-1,1)^{\top},
\bs \mu_3=(1,2)^{\top},
\bs \Sigma_1=\mathrm{diag}(1/10,1),
\bs \Sigma_2=\mathrm{diag}(1/2,1),
\bs \Sigma_3=\mathrm{diag}(1,1)$. 


We are yet studying the nature of deep Gaussian embedding. 
However, as Proposition~\ref{prop:gaussian_not_cpd} shows, 
negative Jeffrey's divergence used in the embedding is already proved to be non-CPD; SIPS cannot approximate it. 
MIPS model is required for approximating such non-CPD kernels. 
Thus we are currently trying to reveal to what extent MIPS applies, by classifying whether each of non-CPD kernels including negative Jeffrey's divergence satisfies the assumption on the kernel $g_*$ in Theorem~\ref{theo:universal_general}.

\section*{Acknowledgement}
We would like to thank Tetsuya Hada for helpful discussions. 
This work was partially supported by JSPS KAKENHI grant 16H02789 to HS and 17J03623 to AO.

\clearpage
\onecolumn

\begin{flushleft}
\textbf{\Large Supplementary Material:} \par
{\Large On a representation power of neural-network based graph embedding and beyond}
\end{flushleft}
\hrulefill

\section{Proofs}

\subsection{Proof of Proposition~\ref{prop:dist_approximation_lower_bound}}

\label{subsec:proof_lower_bound}

With $v=(2M)^{2p}$ and $\int=\int_{[-M,M]^p}$, a lower-bound of 
$\frac{1}{v}
\iint
|
	-\|\bs x-\bs x'\|_2^2
	-
	\langle
		\bs \phi(\bs x),\bs \phi(\bs x')
	\rangle
|
\diff \bs x
\diff \bs x' $ is derived as
\begin{align}
\frac{1}{v}
\iint
\bigg|
	-\|\bs x-\bs x'\|_2^2
	-
	\langle
		\bs \phi(\bs x),\bs \phi(\bs x')
	\rangle
\bigg|
\diff \bs x
\diff \bs x' 
&\geq
\bigg|
\frac{1}{v}
\iint
\left(
	-\|\bs x-\bs x'\|_2^2
	-
	\langle
		\bs \phi(\bs x),\bs \phi(\bs x')
	\rangle
\right)
\diff \bs x
\diff \bs x' \bigg| \nonumber \\
&=
\bigg|
\frac{1}{v}
\iint
\left(
	2\langle \bs x,\bs x' \rangle
	-
	\|\bs x\|_2^2
	-
	\|\bs x'\|_2^2
	-
	\langle
		\bs \phi(\bs x),\bs \phi(\bs x')
	\rangle
\right)
\diff \bs x
\diff \bs x' \bigg| \nonumber \\
&=
\bigg|
\frac{1}{v}
\left(
	2\bigg\| \int \bs x \diff \bs x\bigg\|_2^2
	-
	2 \int \diff \bs x \int \|\bs x\|_2^2 \diff \bs x
	-
	\bigg\|\int \bs \phi(\bs x) \diff \bs x \bigg\|_2^2
\right)
\bigg| \nonumber.
\end{align}
The terms in the last formula are computed as
$\int \bs x \diff \bs x=\bs 0,\int \diff \bs x
=
(2M)^p$,  
\begin{align*}
\int \|\bs x\|_2^2\diff \bs x
&=
\sum_{i=1}^{p} \int x_i^2 \diff \bs x
=
(2M)^{p-1} \sum_{i=1}^{p} \int_{-M}^M x_i^2 \diff x_i
=
(2M)^{p-1}\frac{2pM^3}{3}
=
(2M)^p \frac{pM^2}{3}.
\end{align*}
Considering $\|\int \phi(\bs x)\diff \bs x\|_2^2 \geq 0$, we have
\begin{align*}
\frac{1}{v}
\iint
\bigg|
	-\|\bs x-\bs x'\|_2^2
	-
	\langle
		\bs \phi(\bs x),\bs \phi(\bs x')
	\rangle
\bigg|
\diff \bs x
\diff \bs x' 
\geq
\frac{2}{v}
\int \diff \bs x \int \|\bs x\|_2^2 \diff \bs x
=
\frac{2pM^2}{3}.
\end{align*}
\qed

\subsection{Proof of Theorem~\ref{theo:universal_cpd}}
\label{sec:proof_universal_cpd}
Since $g_*:\mathcal{Y}^2 \to \mathbb{R}$ is a conditionally positive definite kernel on a compact set,
Lemma 2.1 of \citet{berg1984harmonic} indicates that 
$$
	g_{0}(\bs y_*,\bs y_*')
	:=
	g_*(\bs y_*,\bs y_*')
	-
	g_*(\bs y_*,\bs y_0)
	-
	g_*(\bs y_0,\bs y_*')
	+
	g_*(\bs y_0,\bs y_0)
$$
is positive definite for arbitrary $\bs y_0 \in \mathcal{Y}$. 
We fix $\bs y_0$ in the argument below. 
According to \citet{okuno2018probabilistic} Theorem~5.1 (Theorem~\ref{theo:universal_approximate} in this paper), we can specify a neural network $f_{\bs \psi}(\bs x)$ such that
\[
	\sup_{\bs x,\bs x' \in [-M,M]^{p}}
	\bigg|
		g_0\left( f_*(\bs x),f_*(\bs x') \right)
		-
		\langle
			f_{\bs \psi}(\bs x)
			,
			f_{\bs \psi}(\bs x')
		\rangle
	\bigg|
	<
	\varepsilon_1
\]
for any $\varepsilon_1$.
Next, let us consider a continuous function $r(\bs x):=g_*(f_*(\bs x),\bs y_0)-\frac{1}{2}g_*(\bs y_0,\bs y_0)$. 
It follows from the universal approximation theorem \citep{cybenko1989approximation,pmlr-v70-telgarsky17a} that
for any $\varepsilon_2>0$, there exists $T' \in \mathbb{N}$ such that
\[
	\sup_{\bs x \in [-M,M]^{p}} | r(\bs x)  - u_{\bs\xi}(\bs x)  | < \varepsilon_2.
\]
Therefore, we have
\begin{align*}
\sup_{\bs x,\bs x' \in [-M,M]^{p}}
&\biggl|
g_*\left(
	f_*(\bs x)
	,
	f_*(\bs x')
\right)
-
\left\{
\langle f_{\bs \psi}(\bs x) f_{\bs \psi}(\bs x') \rangle
+
u_{\bs\xi}(\bs x)
+
u_{\bs\xi}(\bs x')
\right\}
\biggr|\\
&=
\sup_{\bs x,\bs x' \in [-M,M]^{p}}
\biggl|
\left(
g_0\left(
	f_*(\bs x)
	,
	f_*(\bs x')
\right)
-
\langle f_{\bs \psi}(\bs x) f_{\bs \psi}(\bs x') \rangle
\right) \\
&\hspace{11em}
+
\left(
r(\bs x)
-
u_{\bs\xi}(\bs x)
\right)
+
\left(
r(\bs x')
-
u_{\bs\xi}(\bs x')
\right)
\biggr| \\
&\leq 
\sup_{\bs x,\bs x' \in [-M,M]^{p}}
\biggl|
\left(
g_0\left(
	f_*(\bs x)
	,
	f_*(\bs x')
\right)
-
\langle f_{\bs \psi}(\bs x) f_{\bs \psi}(\bs x') \rangle
\right) \biggr|\\
&\hspace{3em}
+
\sup_{\bs x \in [-M,M]^{p}}
\biggl|
r(\bs x)
-
u_{\bs\xi}(\bs x)
\biggr|
+
\sup_{\bs x' \in [-M,M]^{p}}
\biggl|
r(\bs x')
-
u_{\bs\xi}(\bs x')
\biggr| \\
& <
\varepsilon_1 + 2 \varepsilon_2.
\end{align*}
By letting
$\varepsilon_1=\varepsilon/2,
\varepsilon_2=\varepsilon/4$,
the last formula becomes smaller than $\varepsilon$, thus proving
\[
\sup_{\bs x,\bs x' \in [-M,M]^{p}}
\biggl|
g_*\left(
	f_*(\bs x)
	,
	f_*(\bs x')
\right)
-
\left\{
\langle f_{\bs \psi}(\bs x) f_{\bs \psi}(\bs x') \rangle
+
u_{\bs\xi}(\bs x)
+
u_{\bs\xi}(\bs x')
\right\}
\biggr|
<
\varepsilon.
\]

\qed

\subsection{Proof of Theorem~\ref{theo:universal_cpd2}}
\label{sec:proof_universal_cpd2}
With fixed $\bs y_0 \in \mathcal{Y}$, it follows from \citet{berg1984harmonic} Lemma~2.1 and CPD-ness of the kernel $g_*$ that 
$$
	g_0(\bs y,\bs y')
	:=
	g_*(\bs y,\bs y')
	-
	g_*(\bs y,\bs y_0)
	-
	g_*(\bs y_0,\bs y')
	+
	g_*(\bs y_0,\bs y_0)
$$
is PD.
Since $\mathcal{Y}$ is compact, we have $\sup_{\bs y \in \mathcal{Y}}|g_*(\bs y,\bs y_0)|=a^2$ is bounded.
Let us take a sufficiently large $r>a$ and define  $\tau(\bs y):=\sqrt{r^2 + g_*(\bs y,\bs y_0)}$.
We consider a new kernel
\begin{align*}
g_1(\bs y,\bs y')
:=
g_0(\bs y,\bs y')
+
2\tau(\bs y)\tau(\bs y').
\end{align*}
Since both $g_0(\bs y,\bs y')$ and $\tau(\bs y)\tau(\bs y')$ are PD, $g_1(\bs y,\bs y')$ is also PD. 
Applying Taylor's expansion $\sqrt{1+x}=1+x/2+O(x^2)$, we have
\begin{align*}
\tau(\bs y)
\tau(\bs y')
&=
\sqrt{r^2+g_*(\bs y,\bs y_0)}
\sqrt{r^2+g_*(\bs y',\bs y_0)} \\
&=
r^2
\sqrt{1+g_*(\bs y,\bs y_0)/r^2}
\sqrt{1+g_*(\bs y',\bs y_0)/r^2} \\
&=
r^2
(
	1
	+
	g_*(\bs y,\bs y_0)/2r^2
	+
	O(r^{-4})	
)
(
	1
	+
	g_*(\bs y',\bs y_0)/2r^2
	+
	O(r^{-4})	
) \\
&=
r^2 + \frac{1}{2} (g_*(\bs y,\bs y_0) + g_*(\bs y',\bs y_0))
+
O(r^{-2}),
\end{align*}
thus proving
\begin{align*}
g_1(\bs y,\bs y')
=
g_0(\bs y,\bs y')
+
2\tau(\bs y)\tau(\bs y')
&=
g_*(\bs y,\bs y')
+
g_*(\bs y_0,\bs y_0)
+
2r^2 + O(r^{-2}). 
\end{align*}
Let us define $\gamma:=g_*(\bs y_0,\bs y_0)+2r^2=O(r^2)$. 
Considering the PD-ness of $g_1(\bs y,\bs y') = g_*(\bs y,\bs y') + \gamma+ O(r^{-2})$, we have
\begin{align*}
&\sup_{\bs x,\bs x' \in [-M,M]^p}\bigg|
	g_*(f_*(\bs x),f_*(\bs x'))
	-
	\left(
		\langle f_{\bs \psi}(\bs x),f_{\bs \psi}(\bs x') \rangle
		-
		\gamma
	\right)
\bigg| \\
&=
\sup_{\bs x,\bs x' \in [-M,M]^p}
\bigg|
	g_1(f_*(\bs x),f_*(\bs x'))
	-
	\langle f_{\bs \psi}(\bs x), f_{\bs \psi}(\bs x') \rangle
\bigg|
+
O(r^{-2}) \\
&\leq
\varepsilon + O(r^{-2}).
\end{align*}

\qed

\end{document}